\documentclass[10pt,twocolumn,letterpaper]{article}

\usepackage{cvpr}
\usepackage{times}
\usepackage{epsfig}
\usepackage{graphicx}
\usepackage{amsmath}
\usepackage{amssymb}
\usepackage{multirow}
\usepackage{bm}
\usepackage{subcaption}

\usepackage[pagebackref=true,breaklinks=true,letterpaper=true,colorlinks,bookmarks=false]{hyperref}

\cvprfinalcopy 

\def\cvprPaperID{2013} 

\ifcvprfinal\fi
\begin{document}

\title{Weakly Supervised Action Localization by Sparse Temporal Pooling Network}

%
\author{Phuc Nguyen\thanks{Both authors contributed equally to this work.}\\
University of California\\
Irvine, CA, USA \\
{\tt\small nguyenpx@uci.edu}
\and
Ting Liu\footnotemark[1]\qquad Gautam Prasad\\
Google\\
Venice, CA, USA \\
{\tt\small \{liuti, gautamprasad\}@google.com}
\and
Bohyung Han\\
Seoul National University\\
Seoul, Korea\\
{\tt\small bhhan@snu.ac.kr}
}

\maketitle
\thispagestyle{empty}

\begin{abstract}
We propose a weakly supervised temporal action localization algorithm on untrimmed videos using convolutional neural networks.
Our algorithm learns from video-level class labels and predicts temporal intervals of human actions with no requirement of temporal localization annotations.
We design our network to identify a sparse subset of key segments associated with target actions in a video using an attention module and fuse the key segments through adaptive temporal pooling.
Our loss function is comprised of two terms that minimize the video-level action classification error and enforce the sparsity of the segment selection.
At inference time, we extract and score temporal proposals using temporal class activations and class-agnostic attentions to estimate the time intervals that correspond to target actions.
The proposed algorithm attains state-of-the-art results on the THUMOS14 dataset and outstanding performance on ActivityNet1.3 even with its weak supervision.
\end{abstract}

\section{Introduction}
\label{sec:introduction}
Action recognition and localization in videos are crucial problems for high-level video understanding tasks including, but not limited to, event detection, video summarization, and visual question answering.
Many researchers have been investigating these problems extensively in the last decades, but the main challenge remains the lack of appropriate representation methods of videos.
Contrary to the almost immediate success of convolutional neural networks (CNNs) in many visual recognition tasks for images, applying deep neural networks to videos is not straightforward due to the inherently complex structures of video data, high computation demand, lack of knowledge for modeling temporal information, and so on.
Some attempts to using the representations only from deep learning~\cite{karpathy14large,simonyan14two,tran15learning,wang16temporal} were not significantly better than methods relying on hand-crafted visual features~\cite{laptive05on,wang13action,wang13motionlets}.
As a result, many existing algorithms seek to achieve state-of-the-art performance by combining hand-crafted and learned features.


Many existing video understanding techniques rely on trimmed videos as their inputs. However, most videos in the real world are untrimmed and contain large numbers of irrelevant frames pertaining to target actions and these techniques are prone to fail due to the challenges in extracting salient information. While action localization algorithms are designed to operate on untrimmed videos, they usually require temporal annotations of action intervals, which is prohibitively expensive and time-consuming at large scale. Therefore, it is more practical to develop competitive localization algorithms that require minimal temporal annotations for training.

\begin{figure}[t]
\captionsetup{font=small}
\centering
\includegraphics[width=0.95\linewidth]{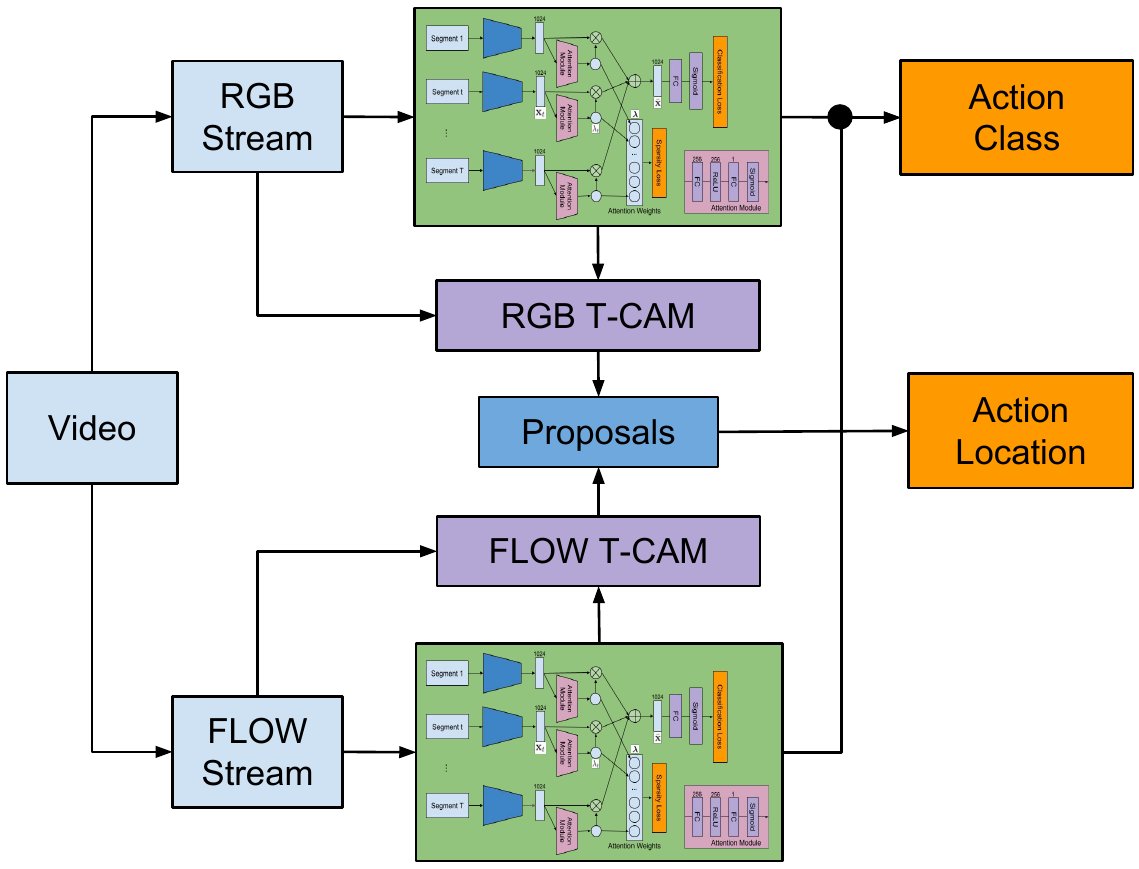}
\caption{Overview of the proposed algorithm. Our algorithm takes a two-stream input---RGB frames and optical flow between frames---from a video, and performs action classification and localization concurrently. For localization, Temporal Class Activation Maps (T-CAMs) are computed from the two streams and employed to generate one dimensional temporal action proposals, from which the target actions are localized in the temporal domain.}
\label{fig:overview}
\end{figure}

Our goal is to temporally localize actions in untrimmed videos.
To this end, we propose a novel deep neural network that learns to select a sparse subset of useful video segments for action recognition in each video by using a loss function that measures the video-level classification error and the sparsity of selected segments.
Temporal Class Activation Maps (T-CAMs) are employed to generate one dimensional temporal proposals used to localize target actions.
Note that we do not exploit temporal annotations of the actions in target datasets during training, and our models are trained only with video-level class labels. 
An overview of our algorithm is shown in Figure~\ref{fig:overview}.

The contributions of this paper are summarized as below.
\begin{itemize}
\item We introduce a principled deep neural network architecture for weakly supervised action localization in untrimmed videos, where actions are detected from a sparse subset of segments identified by the network.
\item We present a method for computing and combining temporal class activation maps and class agnostic attentions for temporal localization of target actions.
\item The proposed weakly supervised action localization technique achieves state-of-the-art results on THUMOS14~\cite{jiang14thumos} and outstanding performance in the ActivityNet1.3~\cite{heilbron15activitynet} action localization task.
\end{itemize}

The rest of this paper is organized as follows.
We discuss the related work in Section~\ref{sec:related} and describe our action localization algorithm in Section~\ref{sec:proposed}.
Section~\ref{sec:experiment} presents the details of our experiment and Section~\ref{sec:conclusion} concludes this paper.

\section{Related Work}
\label{sec:related}


Action recognition aims to identify a single or multiple actions per video and is often formulated as a simple classification problem.
Before the success of CNNs, the algorithm based on improved dense trajectories~\cite{wang13action} presented outstanding performance.
When it comes to the era of deep learning, convolutional neural networks have been widely used.
Afterwards, two-stream networks~\cite{simonyan14two} and 3D convolutional neural networks (C3D)~\cite{tran15learning} are popular solutions to learn video representations and these techniques, including their variations, are extensively used for action recognition.
Recently, a combination of two-stream networks and 3D convolutions, referred to as I3D~\cite{carreira17quo}, was proposed as a generic video representation learning method.
On the other hand, many algorithms develop techniques to recognize actions based on existing representation methods~\cite{wang16temporal,wang17spatiotemporal,feichtenhofer16spatiotemporal,girdhar17actionvlad,feichtenhofer17spatiotemporal,shi17learning}.

Action localization is different from action recognition, because it requires the detections of temporal or spatiotemporal volumes containing target actions.
There are various existing methods based on deep learning including structured segment network~\cite{zhao17temporal}, contextual relation learning~\cite{soomro15action}, multi-stage CNNs~\cite{shou16temporal}, temporal association of frame-level action detections~\cite{gkioxari15finding}, and techniques using recurrent neural networks~\cite{yeung16end,ma16learning}.
Most of these approaches rely on supervised learning and employ temporal or spatio-temporal annotations to train the models.
To facilitate action detection and localization, many algorithms use action proposals~\cite{buch17sst,escorcia16daps,wang16actioness}, which is an extension of object proposals for object detection in images.

There are only a few approaches based on weakly supervised learning that rely solely on video-level class labels to localize actions in temporal domain.
UntrimmedNet~\cite{wang17untrimmednets} learns attention weights on precut video segments using a temporal softmax function and thresholds the attention weights to generate action proposals. The algorithm improves the video-level classification performance.
However, generating action proposals solely from class-agnostic attention weights is suboptimal and the use of the softmax function across proposals may not be effective to detect multiple instances.
Hide-and-seek~\cite{singh17hide} proposes a technique that randomly hides regions to force residual attention learning and thresholds class activation maps at inference time for weakly supervised spatial object detection and temporal action localization.
While working well at spatial localization tasks, this method fails to show satisfactory performance in temporal action localization tasks in videos.
Both algorithms are motivated by the recent success of weakly supervised object localization in images.
In particular, the formulation of UntrimmedNet for action localization heavily relies on the idea proposed in \cite{bilen16weakly}.

There are some other approaches~\cite{bojanowski14weakly,huang16connectionist,rechard17weakly} that learn to localize or segment actions in a weakly supervised setting by exploiting the temporal order of subactions during training.
The main objective of these studies is to find the boundaries of sequentially presented subactions, while our approach aims to extract temporal intervals of full actions from input videos.

There are several publicly available datasets for action recognition including UCF101~\cite{soomro12ucf101}, Sports-1M~\cite{karpathy14large}, HMDB51~\cite{kuehne11hmdb}, Kinetics~\cite{kay2017kinetics} and AVA~\cite{gu17ava}.
The videos in these datasets are trimmed so that the target actions appear throughout each clip.
In contrast, THUMOS14 dataset~\cite{jiang14thumos} and ActivityNet~\cite{heilbron15activitynet} provide untrimmed videos that contain background frames and temporal annotations about which frames are relevant to the target actions.
Note that each video in THUMOS14 and ActivityNet may have multiple actions happening in a single frame.


%
\begin{figure*}[t]
\captionsetup{font=small}
\centering
\includegraphics[width=0.86\linewidth]{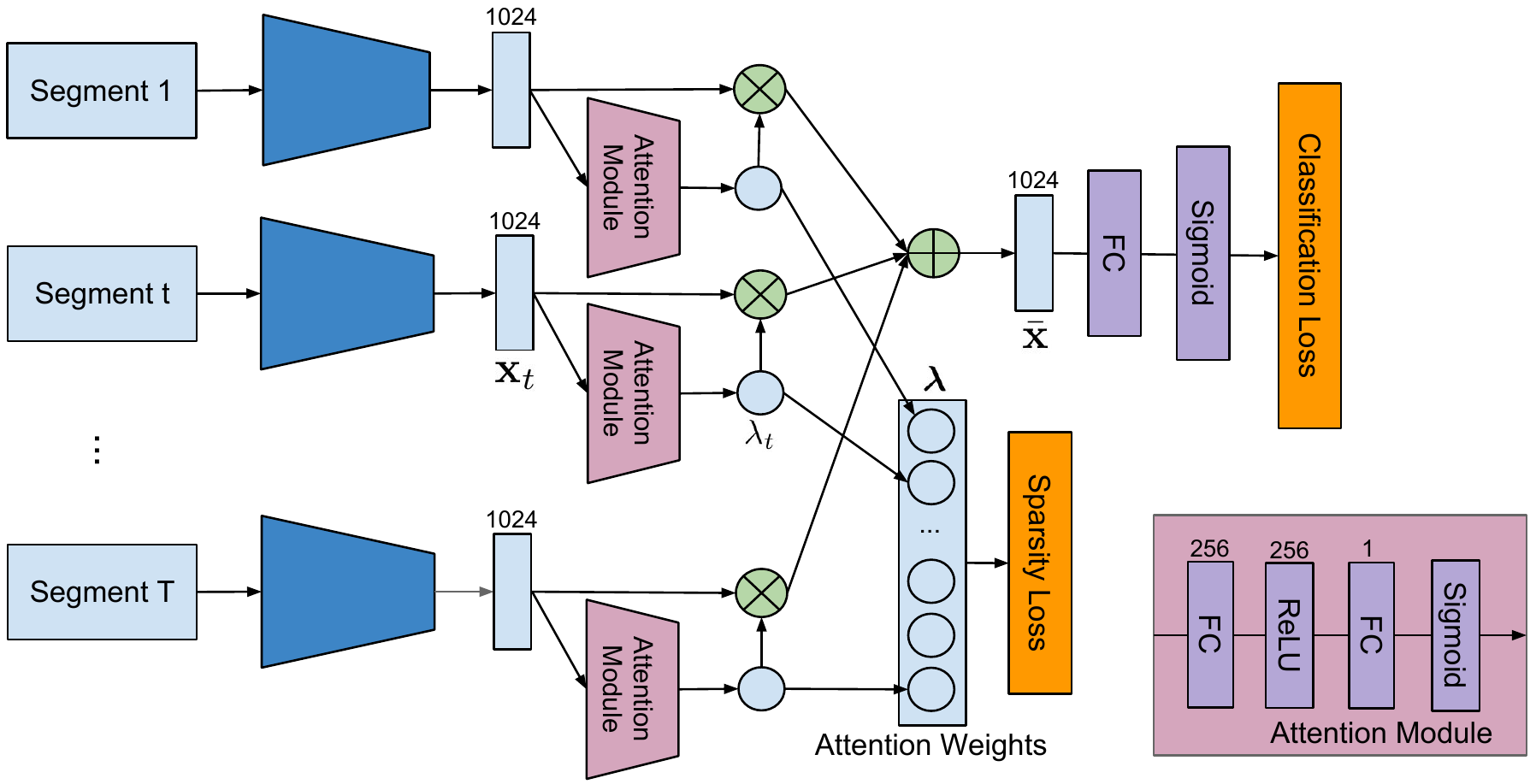}
	\caption{Network architecture for our weakly supervised temporal action localization model.  We first extract feature representations for a set of uniformly sampled video segments using a pretrained network. The attention module computes class-agnostic attention weights for each segment, which are used to generate a video-level representation via weighted temporal average pooling. The representation is given to the classification module that can be trained with regular cross entropy loss with video-level labels. An $\ell_1$ loss is placed on the attention weights to enforce sparse attentions.}
\label{fig:architecture}
\end{figure*}

\section{Proposed Algorithm}
\label{sec:proposed}

%
We claim that an action can be recognized from a video by identifying a set of key segments presenting important action components. So we design a neural network that learns to measure the importance of each segment in a video and automatically selects a sparse subset of representative segments to predict the video-level class labels. Only ground-truth video-level class labels are required for training the model.
For action localization at inference time, we first identify relevant classes in each video and then generate temporal action proposals from temporal class activations and attentions to find the temporal location of each relevant class.
The network architecture for our weakly supervised action recognition component is illustrated in Figure~\ref{fig:architecture}.
We describe each step of our algorithm in the rest of this section.

\subsection{Action Classification}
\label{sub:weakly}
To predict class labels in each video, we sample a set of segments and extract feature representations from each segment using pretrained convolutional neural networks.
Each feature vector is then fed to an attention module that consists of two fully connected (FC) layers and a ReLU layer located between the two FC layers.
The output of the second FC layer is given to a sigmoid function that enforces the generated attention weights to be between 0 and 1.
These class-agnostic attention weights are then used to modulate the temporal average pooling---a weighted sum of the feature vectors---to create a video-level representation.
We pass this representation through an FC layer followed by a sigmoid layer to obtain class scores.

Formally, let ${\bf x}_t \in \mathbb{R}^m$ be the $m$ dimensional feature representation extracted from a video segment centered at time $t$, and $\lambda_t$ be the corresponding attention weight. 
The video level representation, denoted by $\bar{{\bf x}}$, corresponds to an attention weighted temporal average pooling, which is given by
\begin{equation}
\bar{{\bf x}} = \sum_{t=1}^T \lambda_t {\bf x}_t,
\label{eq:video_feature}
\end{equation}
where ${\bm \lambda} = (\lambda_1, \dots, \lambda_T)^{\top}$ is a vector of scalar outputs from the attention module and $T$ is the total number of sampled video segments.
The attention weight vector ${\bm \lambda}$ is defined in a class-agnostic way, which is useful to identify segments relevant to all the actions of interest and estimate the temporal intervals of the detected actions.

The loss function in the proposed network is composed of two terms, the classification loss and the sparsity loss, which is given by%
\begin{equation}
\mathcal{L} = \mathcal{L}_\text{class} + \beta \cdot \mathcal{L}_\text{sparsity}, 
\label{eq:loss}
\end{equation}
where $\mathcal{L}_\text{class}$ denotes the classification loss computed on the video-level class labels, $\mathcal{L}_\text{sparsity}$ is the sparsity loss on the attention weights, and $\beta$ is a constant to control the trade-off between the two terms.
The classification loss is based on the standard multi-label cross-entropy loss between ground-truth and $\bar{{\bf x}}$ (after passing through a few layers as illustrated in Figure~\ref{fig:architecture}), while the sparsity loss is given by the $\ell_1$ norm on attention weights $|| {\bm \lambda} ||_1$.
Because of the use of the sigmoid function and the $\ell_1$ loss, all the attention weights tend to have values close to either 0 or 1.
Note that integrating the sparsity loss is aligned with our claim that an action can be recognized with a sparse subset of key segments in a video.


\subsection{Temporal Class Activation Mapping}
\label{sub:temporal}
\begin{figure*}[ht]
\captionsetup{font=small}
\centering
\includegraphics[width=1\linewidth]{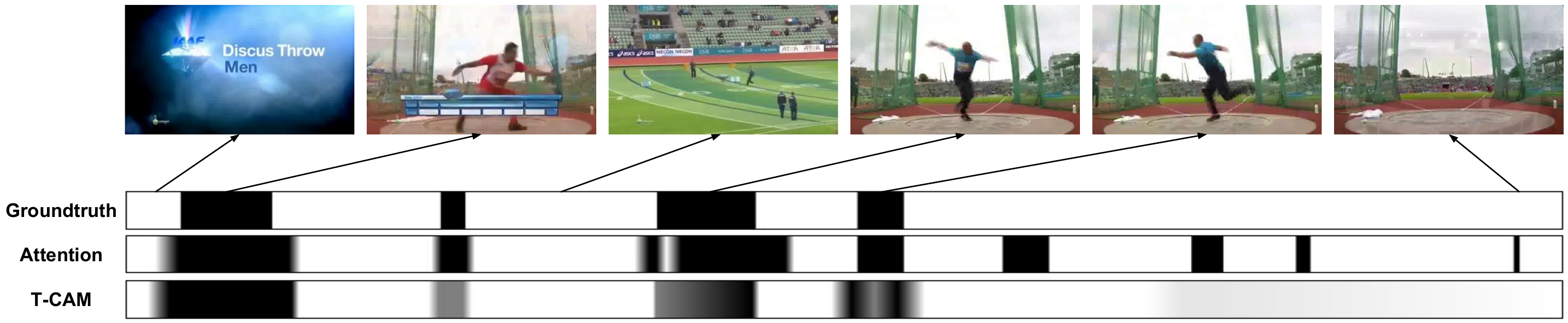}
    \caption{Illustration of the ground-truth temporal intervals for the {\it ThrowDiscus} class, the temporal attentions, and the T-CAM for an example video in the THUMOS14 dataset~\cite{jiang14thumos}. The horizontal axis in the plots denote the timestamps. In this example, the T-CAM values for {\it ThrowDiscus} provide accurate action localization information.  Note that the temporal attention weights are large at several locations that do not correspond to the ground-truth annotations.  This is because temporal attention weights are trained in a class-agnostic way.}
\label{fig:tcam_example}
\end{figure*}
To identify the time intervals corresponding to target actions, we extract a number of action interval candidates.
Based on the idea in \cite{zhou16learning}, we derive a one dimensional class-specific activation map in the temporal domain, referred to as the Temporal Class Activation Map (T-CAM).
Let ${\bf w}^{c}(k)$ denote the $k$-th element in the weight parameter ${\bf w}^c$ of the final fully connected layer, where the superscript $c$ represents the index of a particular class.
The input to the final sigmoid layer for class $c$ is
\begin{align}
s^c & = \sum_{k=1}^m {\bf w}^{c}(k) \bar{\bf x}(k) \nonumber \\
 & = \sum_{k=1}^m {\bf w}^{c}(k) \sum_{t=1}^T \lambda_t {\bf x}_t(k) \\
 & = \sum_{t=1}^T \lambda_t \sum_{k=1}^m {\bf w}^{c}(k) {\bf x}_t(k). \nonumber
\end{align}

T-CAM, denoted by ${\bf a}_t = (a_t^1, a_t^2, \dots, a_t^{C})^{\top}$, indicates the relevance of the representations to each class at time step $t$, where each element $a_t^c$ for class $c$ ($c = 1, \dots, C$) is given by
\begin{equation}
a_t^c = \sum_{k=1}^m {\bf w}^c(k) {\bf x}_t(k).
\label{eq:t-cam}
\end{equation}
%

Figure~\ref{fig:tcam_example} illustrates an example of the attention weights and the T-CAM outputs in a video given by the proposed algorithm.
We can observe that the discriminative temporal regions are effectively highlighted by the attention weights and the T-CAMs.
Also, some temporal intervals with large attention weights do not correspond to large T-CAM values because such intervals may represent other actions of interest.
The attention weights measure the generic actionness of temporal video segments while the T-CAMs present class-specific information.

\subsection{Two-stream CNN Models} 
\label{sub:two-stream}
We employ the recently proposed I3D model~\cite{carreira17quo} to compute feature representations for the sampled video segments.
Using multiple streams of information such as RGB and optical flow has become a standard practice in action recognition and detection~\cite{carreira17quo,feichtenhofer16convolutional,simonyan14two} as it often provides a significant boost in performance.
We also train two action recognition networks separately with identical settings as illustrated in Figure~\ref{fig:architecture} for the RGB and the flow stream.
Note that our I3D networks are pretrained on the Kinetics dataset~\cite{kay2017kinetics}, and we only use it as feature extraction machines without any fine-tuning on our target datasets.
Our two-stream networks are then fused to localize actions in an input video.
The procedure is discussed in the following subsection.

\subsection{Temporal Action Localization}
\label{sub:temporal}
For an input video, we identify relevant class labels based on video-level classification scores (Section~\ref{sub:weakly}).
For each relevant action, we generate temporal proposals, \ie, one-dimensional time intervals, with their class-specific confidence scores, corresponding to segments that potentially enclose the target actions.

To generate temporal proposals, we compute the T-CAMs for both the RGB and the flow streams, denoted by $a_{t,\text{RGB}}^c$ and $a_{t,\text{FLOW}}^c$ respectively, based on \eqref{eq:t-cam} and use them to derive the weighted T-CAMs, $\psi^c_\text{t, \text{RGB}}$ and $\psi^c_\text{t, \text{FLOW}}$ as
\begin{align}
\psi^c_{t,\text{RGB}} & = \lambda_{t, \text{RGB}} \cdot \text{sigmoid}(a_{t, \text{RGB}}^c) \\
\psi^c_{t, \text{FLOW}} & = \lambda_{t, \text{FLOW}} \cdot \text{sigmoid}(a_{t, \text{FLOW}}^c).
\end{align}
Note that $\lambda_t$ is an element of the sparse vector $\bm \lambda$, and multiplying $\lambda_t$ can be interpreted as a soft selection of the values from the following sigmoid function.
Similar to~\cite{zhou16learning}, we threshold the weighted T-CAMs, $\psi^c_{t,\text{RGB}}$ and $\psi^c_{t,\text{FLOW}}$ to segment these signals. The temporal proposals are then the one-dimensional connected components extracted from each stream.
It is intuitive to generate action proposals using the weighted T-CAMs, instead of directly from the attention weights, because each proposal should contain a single kind of action.
Optionally, we linearly interpolate the weighted T-CAM signals between sampled segments before thresholding to improve the temporal resolution of the proposals with minimal computation addition.

Unlike the original CAM-based bounding box proposals~\cite{zhou16learning} where only the largest bounding box is retained, we keep all the connected components that pass the predefined threshold. 
Each proposal $[t_{\text{start}},t_{\text{end}}]$ is assigned a score for each class $c$, which is given by the weighted average T-CAM of all the frames within the proposal:
%
%
\begin{equation}
   \sum_{t=t_{\text{start}}}^{t_{\text{end}}}\lambda_{t, *} \frac{\alpha \cdot a^c_{t,\text{RGB}}+ (1-\alpha) \cdot a^c_{t,\text{FLOW}}}{t_{\text{end}}-t_{\text{start}}+1}\label{eq:box_score},
\end{equation}%
where $* \in \{ \text{RGB}, \text{FLOW} \}$ and $\alpha$ is a parameter to control the magnitudes of the two modality signals.
Finally, we perform non-maximum suppression among temporal proposals of each class independently to remove highly overlapped detections.

\subsection{Discussion}
\label{sub:discussion}

Our algorithm attempts to localize actions in untrimmed videos temporally by estimating sparse attention weights and T-CAMs for generic and specific actions, respectively.
The proposed method is principled and novel when compared to the existing UntrimmedNet~\cite{wang17untrimmednets} because of the following reasons.
\begin{itemize}
\item Our model has a unique deep neural network architecture with classification and sparsity losses.
\item Our action localization procedure is based on a completely different pipeline that leverages class-specific action proposals using T-CAMs.
\end{itemize}
%
Note that~\cite{wang17untrimmednets} follows a similar framework used in~\cite{bilen16weakly}, where softmax functions are employed across both action classes and proposals; it has a critical limitation in handling multiple action classes and instances in a single video.


Similar to pretraining on the ImageNet dataset~\cite{deng09imagenet} for weakly supervised learning problems in images, we utilize features from I3D models~\cite{carreira17quo} pretrained on the Kinetics dataset~\cite{kay2017kinetics} for video representation. 
Although the Kinetics dataset has considerable class overlap with our target datasets, its video clips are mostly short and contain only parts of actions, which makes their characteristics different from the ones in our untrimmed target datasets. 
We also do not fine-tune the I3D models and our network may not be optimized for the classes in the target tasks and datasets.


%
\begin{table*}[t]
\captionsetup{font=small}
\centering
\caption{Comparison of our algorithm with other recent techniques on the THUMOS14 testing set. We divide the algorithms into two groups depending on their levels of supervision. Each group is sorted chronologically, from older to newer ones. STPN, including the version using UntrimmedNet features, clearly presents state-of-the-art performance in the weakly supervised setting and is even competitive with many fully supervised approaches. 
}
\label{table:comparison_thumos14}
\small
\scalebox{0.95}{
\begin{tabular}{c|c||ccccccccc}
\multirow{2}{*}{Supervision} & \multirow{2}{*}{Method} & \multicolumn{9}{c}{AP@IoU}  \\
&  & 0.1 & 0.2 & 0.3 & 0.4 & 0.5 & 0.6 & 0.7 & 0.8 & 0.9 \\
\hline
\multirow{11}{*}{\shortstack{Fully \\ supervised}} & Heilbron~\etal.~\cite{heilbron16fast} & -- & -- & -- & -- & 13.5  & -- & -- & -- & -- \\
& Richard~\etal~\cite{richard16temporal}  & 39.7 & 35.7 & 30.0 & 23.2 & 15.2 & --& -- & -- & -- \\
& Shou~\etal~\cite{shou16temporal}  	& 47.7 & 43.5 & 36.3 & 28.7 & 19.0 & 10.3 & {\color{white}0}5.3 & -- & -- \\
& Yeung~\etal~\cite{yeung16end}  	& 48.9 & 44.0 & 36.0 & 26.4 & 17.1 & -- & -- & -- & -- \\
& Yuan~\etal~\cite{yuan16temporal}  	& 51.4 & 42.6 & 33.6 & 26.1 & 18.8 & -- & -- & -- & -- \\
& Escorcia~\etal~\cite{escorcia16daps}	& -- & -- & -- & -- & 13.9 & -- & -- & -- & -- \\
& Shou~\etal~\cite{shou17cdc}		& -- & -- & 40.1 & 29.4 & 23.3 & 13.1 & {\color{white}0}7.9 & -- & -- \\
& Yuan~\etal\cite{yuan17temporal}	& 51.0 & 45.2 & 36.5 & 27.8 & 17.8 & -- & -- & -- & -- \\
& Xu~\etal\cite{xu17r}		& 54.5 & 51.5 & 44.8 & 35.6 & 28.9 & -- & -- & -- & -- \\
& Zhao~\etal~\cite{zhao17temporal}		& 66.0 & 59.4 & 51.9 & 41.0 & 29.8 & -- & -- & -- & -- \\
& Alwasssel~\etal~\cite{alwassel17action} & 49.6 & 44.3 & 38.1 & 28.4 & 19.8 & -- & -- & -- & -- \\
\hline
\multirow{4}{*}{\shortstack{Weakly \\ supervised}} & Wang~\etal~\cite{wang17untrimmednets} & 44.4 & 37.7 & 28.2 & 21.1 & 13.7 & -- & -- & -- & -- \\
& Singh \& Lee~\cite{singh17hide} & 36.4 & 27.8 & 19.5 & 12.7 & {\color{white}0}6.8 & -- & -- & -- & -- \\ \cline{2-11}
& STPN	   & 52.0 & 44.7 & 35.5 & 25.8 & 16.9 & {\color{white}0}9.9 & {\color{white}0}4.3 & {\color{white}0}1.2 & {\color{white}0}0.1 \\
& STPN with UntrimmedNet features & 45.3 & 38.8 & 31.1 & 23.5 & 16.2 & {\color{white}0}9.8 & {\color{white}0}5.1 & {\color{white}0}2.0 & {\color{white}0}0.3 \\
\hline
\end{tabular}}
\end{table*}

\section{Experiments}
\label{sec:experiment}

This section first describes the details of the benchmark datasets and the evaluation setup.
Our algorithm, referred to as Sparse Temporal Pooling Network (STPN), is compared with other state-of-the-art techniques based on fully and weakly supervised learning.
Finally, we analyze the contribution of individual components in our algorithm.

\subsection{Datasets and Evaluation Method}
\label{sub:datasets}
We evaluate STPN on two popular action localization benchmark datasets, THUMOS14~\cite{jiang14thumos} and ActivityNet1.3~\cite{heilbron15activitynet}.
Both datasets are untrimmed, meaning the videos include frames that contain no target actions and we do not exploit the temporal annotations for training.
Note that there may exist multiple actions in a single video and even in a single frame in these datasets.

The THUMOS14 dataset has video-level annotations of 101 action classes in its training, validation, and testing sets, and temporal annotations for a subset of videos in the validation and testing sets for 20 classes.
We train our model with the 20-class validation subset, which consists of 200 untrimmed videos, without using the temporal annotations.
We evaluate our algorithm using the 212 videos in the 20-class testing subset with temporal annotations.
This dataset is challenging as some videos are relatively long (up to 26 minutes) and contain multiple action instances.
The length of an action varies significantly, from less than a second to minutes.

The ActivityNet dataset is a recently introduced benchmark for action recognition and localization in untrimmed videos.
We use ActivityNet1.3, which originally consisted of 10,024 videos for training, 4,926 for validation, and 5,044 for testing\footnote{In our experiments, there were 9740, 4791, and 4911 videos accessible from YouTube in the training, validation, and testing set respectively.}, with 200 activity classes.
This dataset contains a large number of natural videos that involve various human activities under a semantic taxonomy.

We follow the standard evaluation protocol based on mean average precision (mAP) values at several different levels of intersection over union (IoU) thresholds.
The evaluation of both the datasets is conducted using the benchmarking code for the temporal action localization task provided by ActivityNet\footnote{\url{https://github.com/activitynet/ActivityNet/blob/master/Evaluation/}}.
The result on the ActivityNet1.3 testing set is obtained by submitting results to the evaluation server.

\subsection{Implementation Details}
We use two-stream I3D networks~\cite{carreira17quo} trained on the Kinetics dataset~\cite{kay2017kinetics} to extract features for video segments.
For the RGB stream, we rescale the smallest dimension of a frame to $256$ and perform the center crop of size $224 \times 224$.
For the flow stream, we apply the TV-$L1$ optical flow algorithm~\cite{wedel09improved}.
The inputs to the I3D models are stacks of $16$ (RGB or flow) frames sampled at $10$ frames per second.

We sample $400$ segments at uniform interval from each video in both training and testing.
During training, we perform stratified random perturbation on the segments sampled for data augmentation. 
The network is trained using Adam optimizer with learning rate $10^{-4}$.
At testing time, we first reject classes whose video-level probabilities are below $0.1$, and then retrieve one-dimensional temporal proposals for the remaining classes.
We set the modality balance parameter $\alpha$ in \eqref{eq:box_score} to $0.5$.
Our algorithm is implemented in TensorFlow.

\subsection{Results}
\label{sub:results}
Table~\ref{table:comparison_thumos14} summarizes the test results on THUMOS14 for action localization methods in the past two years.
We included both fully and weakly supervised approaches in the table.
Our algorithm outperforms the other two existing approaches based on weakly supervised learning~\cite{wang17untrimmednets,singh17hide}. 
Even with significant difference in the level of supervision, our algorithm presents competitive performance to several recent fully supervised approaches.
We also present performance of our model using the features extracted from the pretrained UntrimmedNet~\cite{wang17untrimmednets} two-stream models to evaluate the performance of our algorithm based on weakly supervised representation learning. 
For this experiment, we adjust $\alpha$ to $0.1$ to handle the heterogeneous signal magnitudes of the two modalities.
From~Table~\ref{table:comparison_thumos14}, we can see that STPN also outperforms the UntrimmedNet~\cite{wang17untrimmednets} and the Hide-and-Seek algorithm~\cite{singh17hide} in this setting.

We also present performance of our algorithm on the validation and the testing set of ActivityNet1.3 dataset in Table~\ref{table:comparison_activitynet_validation} and \ref{table:comparison_activitynet_test}, respectively.
We can see that our algorithm outperforms some fully supervised approaches on both the validation and the testing set.
Note that most of the action localization results available on the leaderboard are specifically tuned for the ActivityNet Challenge, which may not be directly comparable with our algorithm.
To our knowledge, this is the first attempt to evaluate weakly supervised action localization performance on this dataset, and we report the results as a baseline for future reference.

\begin{figure*}[t]
\captionsetup{font=small}
\centering
	\begin{subfigure}[b]{\textwidth}
		\includegraphics[width=\textwidth]{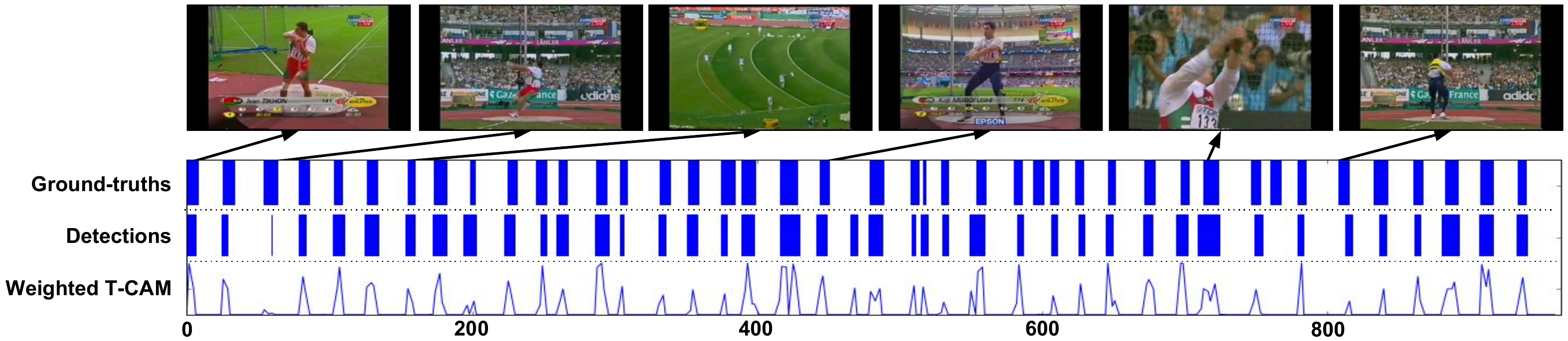}
		\caption{An example of the {\it HammerThrow} action.}
		\label{fig:qualitative_impressive}
    \end{subfigure}
    \begin{subfigure}[b]{\textwidth}
       	\includegraphics[width=\textwidth]{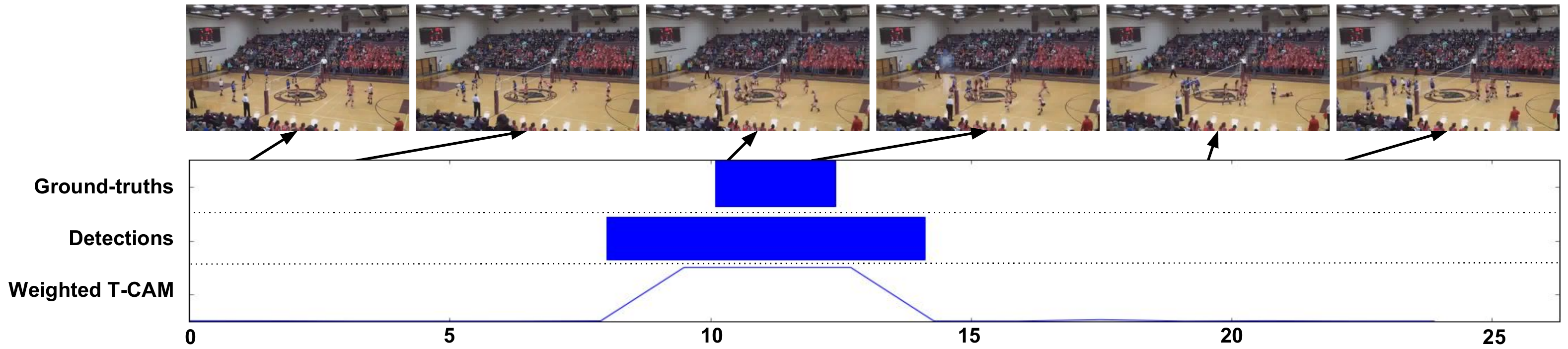}
       	\caption{An example of the {\it VolleyballSpiking} action.}
       	\label{fig:qualitative_impressive_2}
        \end{subfigure}
    \begin{subfigure}[b]{\textwidth}
    	\includegraphics[width=\textwidth]{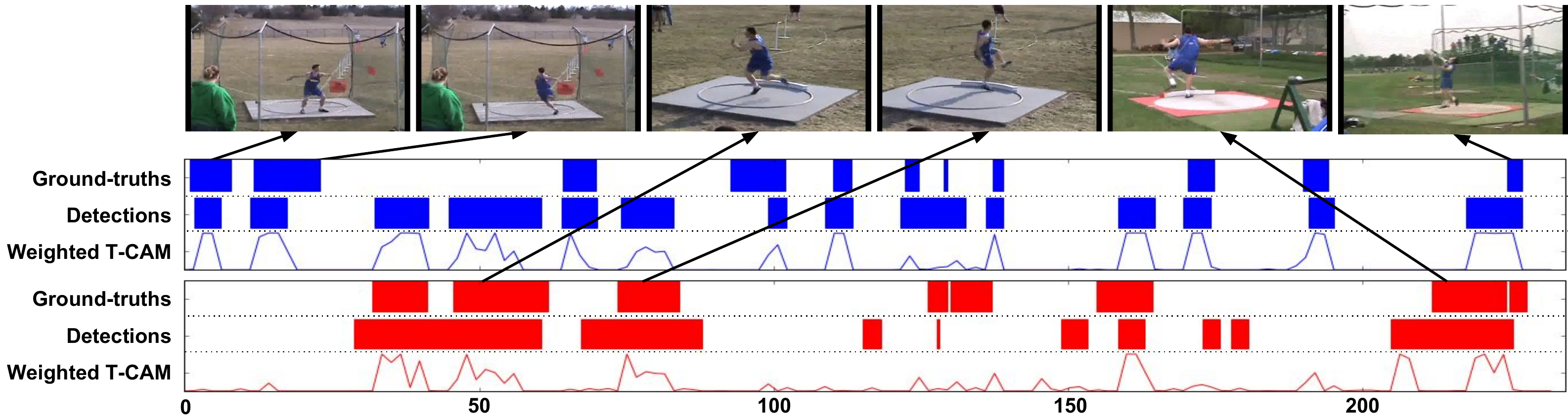}
    	\caption{An example of the {\it ThrowDiscus} (blue) and {\it Shotput} (red) actions.}
   	\label{fig:qualitative_multi}
    \end{subfigure}
    \begin{subfigure}[b]{\textwidth}
        \includegraphics[width=\textwidth]{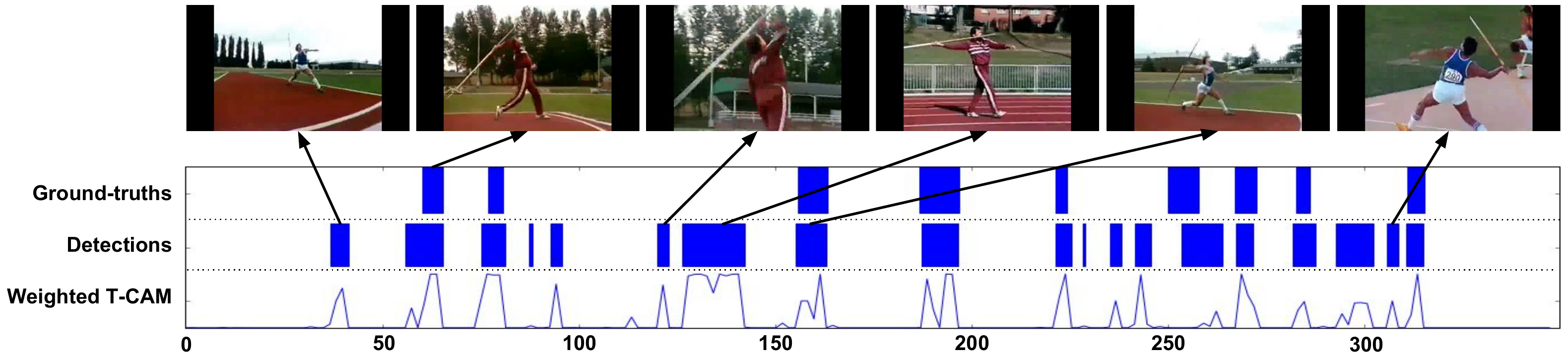}
        \caption{An example of the {\it JavelinThrow} action.}
        \label{fig:qualitative_failure}
    \end{subfigure}
    \caption{Qualitative results on THUMOS14. The horizontal axis in the plots denote the timestamps (in seconds). (a) There are many action instances in the input videos and our algorithm shows good action localization performance. (b) The appearance of the video remains similar from the beginning to the end. There is little motion between frames. Our model is still able to localize the time window where the action actually happens. (c) Two different actions appear in a single video and their appearance and the motion patterns are similar. Even in the case, the proposed algorithm successfully identifies two actions accurately despite some false positives.  (d) Our results have several false positives, but they are often from missing ground-truth annotations. Another source of false alarms is the similarity of the observed actions to the target action.}
\label{fig:qualitative_results}
\end{figure*}
Figure~\ref{fig:qualitative_results} demonstrates qualitative results on the THUMOS14 dataset.
As mentioned in Section~\ref{sub:datasets}, videos in this dataset are often long and contain many action instances, which may be composed of multiple categories.
Figure~\ref{fig:qualitative_impressive} presents an example with a number of action instances along with our predictions and the corresponding T-CAM signals.
Our algorithm effectively pinpoints the temporal boundaries of many action instances.
In Figure~\ref{fig:qualitative_impressive_2}, the appearance of all the frames are similar, and there is little motion between frames.
Despite these challenges, our model still localizes the target action fairly well.
Figure~\ref{fig:qualitative_multi} illustrates an example of a video containing action instances from two different classes. 
Visually, the two involved action classes---{\it Shotput} and {\it ThrowDiscus}---are similar in their appearance (green grass, person with blue shirt, on a gray platform) and motion patterns (circular throwing).
STPN is able to not only localize the target actions but also classify the action categories successfully, despite several short-term false positives.
Figure~\ref{fig:qualitative_failure} shows a instructional video for {\it JavelinThrow}, where our algorithm detects most of the ground-truth action instances while it also generates many false positives. 
There are two causes for the false alarms.
First, the ground-truth annotations for {\it JavelinThrow} are often missing, making true detections counted as false positives.
The second source is related to the segments, where the instructors demonstrate javelin throwing but only parts of such actions are visible.
These segments resemble a real {\it JavelinThrow} action in both appearance and motion.
\begin{table}[t]
\captionsetup{font=small}
\centering
\caption{Results on the ActivityNet1.3 validation set. The entries with an asterisk (*) are from the ActivityNet Challenge submissions. Note that~\cite{shou17cdc} is the result of post-processing based on~\cite{wang16anet}, making the comparison difficult.}
\label{table:comparison_activitynet_validation}
\vspace{-0.2cm}
\small
\scalebox{0.9}{
\begin{tabular}{c|c||ccc}
\multirow{2}{*}{} & \multirow{2}{*}{Method} & \multicolumn{3}{c}{AP@IoU}  \\
& & 0.5 & 0.75 & 0.95 \\
\hline
\multirow{6}{*}{\shortstack{Fully \\ supervised}} & Singh \& Cuzzolin~\cite{singh16untrimmed}*	& 34.5 & -- & -- \\
& Wang \& Tao~\cite{wang16anet}* & 45.1 & {\color{white}0}4.1 & {\color{white}0}0.0 \\
& Shou~\etal~\cite{shou17cdc}* & 45.3 & 26.0 & {\color{white}0}0.2 \\
& Xiong~\etal~\cite{xiong17pursuit}* & 39.1 & 23.5 & {\color{white}0}5.5 \\
\cline{2-5}
& Montes~\etal~\cite{montes16temporal}		& 22.5 & -- & -- \\
& Xu~\etal~\cite{xu17r}	& 26.8 & -- & -- \\
\hline
\shortstack{Weakly \\ supervised} & \shortstack{ \vspace{0.15cm} \\ STPN \\  \vspace{0.1cm}} & \shortstack{ \vspace{0.15cm} \\ 29.3 \\  \vspace{0.1cm}}  & \shortstack{ \vspace{0.15cm} \\ 16.9 \\  \vspace{0.1cm}}  & \shortstack{ \vspace{0.15cm} \\ {\color{white}0}2.6 \\  \vspace{0.1cm}} \\
\hline
\end{tabular}
}
\end{table}
\begin{table}[t]
\captionsetup{font=small}
\centering
\caption{Results on the ActivityNet1.3 testing set. The entries with an asterisk (*) are from the ActivityNet Challenge submissions.}
\label{table:comparison_activitynet_test}
\vspace{-0.2cm}
\small
\scalebox{0.9}{
\begin{tabular}{c|c||c}
 & Method & mAP\\
\hline
\multirow{5}{*}{\shortstack{Fully \\ supervised}} & Singh \& Cuzzolin~\cite{singh16untrimmed}* & 17.83 \\
& Wang \& Tao~\cite{wang16anet}* & 14.62\\
& Xiong~\etal~\cite{xiong17pursuit}* & 26.05 \\
\cline{2-3}
& Singh~\etal~\cite{singh16multi} & 17.68\\
& Zhao~\etal~\cite{zhao17temporal} & 28.28\\
\hline
\shortstack{Weakly \\ supervised} &  \shortstack{ \vspace{0.15cm} \\ STPN \\  \vspace{0.1cm}}	 	&  \shortstack{ \vspace{0.15cm} \\ 20.07 \\  \vspace{0.1cm}} \\
\hline
\end{tabular}
}
\end{table}

\subsection{Ablation Study}
\label{sub:ablation}
We investigate the contribution of several components proposed in our weakly supervised architecture and implementation variations.
All the experiments in our ablation study are performed on the THUMOS14 dataset.

\begin{figure}[t]
\captionsetup{font=small}
\centering
\includegraphics[width=0.9\linewidth]{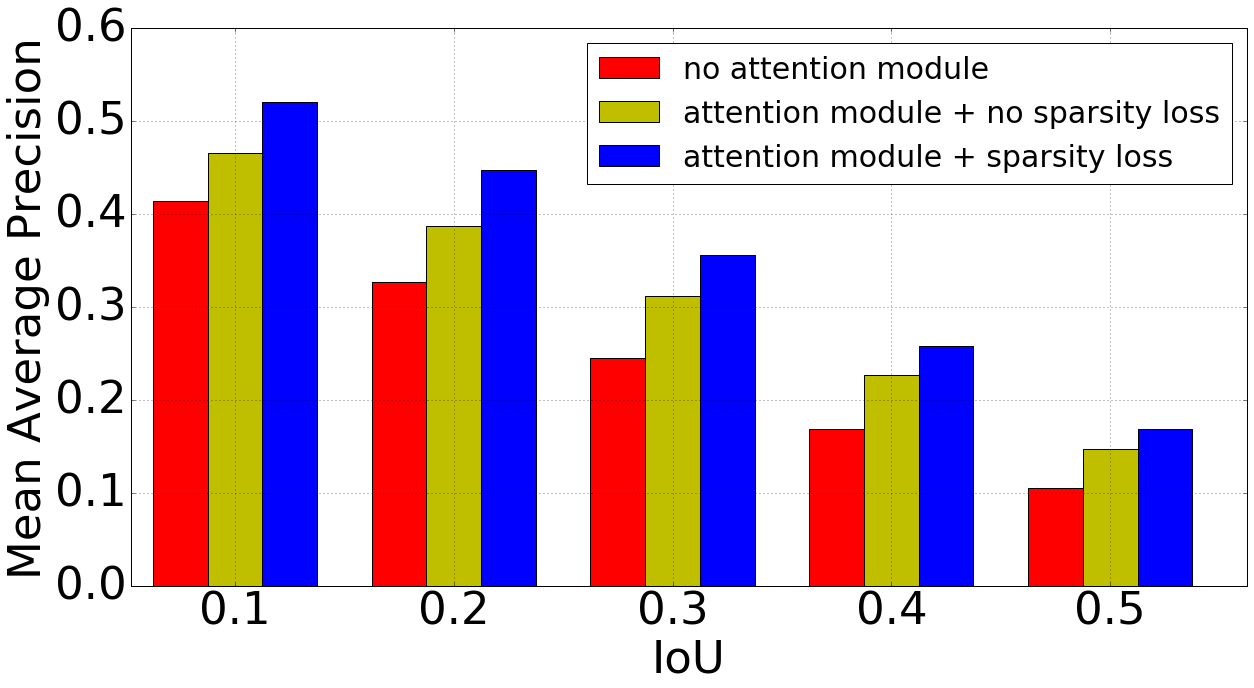}\\
\vspace{-0.1cm}
    \caption{Performance with respect to architectural variations. The attention module is useful as it allows the model to explicitly focus on important parts of input videos.  Enforcing sparsity in action recognition via $\ell_1$ loss gives significant boost to the performance.}
\label{fig:experiment_architecture_choice}
\end{figure}

\paragraph{Choice of architectures}
Our premise is that an action can be recognized with a sparse subset of segments in a video. When we learn our action classification network, two loss terms---classification and sparsity losses---are employed.
Our baseline is the architecture without the attention module and the sparsity loss, which share the motivation with the architecture in~\cite{zhou16learning}. 
We also test another baseline with the attention module but without the sparsity loss.
Figure~\ref{fig:experiment_architecture_choice} shows the comparisons between our baselines and the full model. 
We observe that both the sparsity loss and the attention weighted pooling make substantial contributions to the performance improvement.

\begin{figure}[t]
\captionsetup{font=small}
\centering
\includegraphics[width=0.9\linewidth]{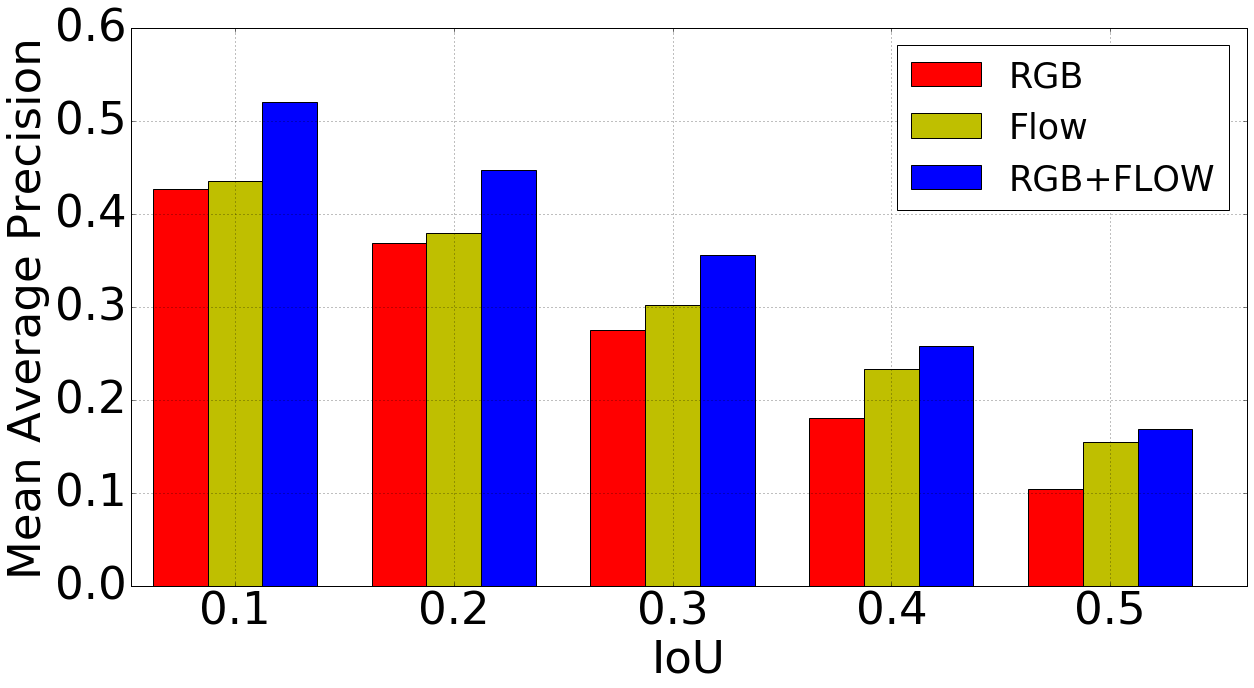}\\
\vspace{-0.1cm}
    \caption{Performance with respect to modality choices. Optical flow offers stronger cues than the RGB frames for action localization and the combination of the two features leads to significant performance improvement.}
\label{fig:experiment_feature_choice}
\end{figure}

\paragraph{Choice of modalities} 
As mentioned in Section~\ref{sub:two-stream}, we use two-stream I3D networks for generating temporal action proposals and computing the attention weights. We also combine the two modalities for scoring the proposals.
Figure~\ref{fig:experiment_feature_choice} illustrates the effectiveness of each modality and their combination. 
When comparing the individual performance of each modality, the flow stream offers stronger performance than the RGB steam.
Similar to action recognition, the combination of these modalities provides significant performance improvement.

\section{Conclusion}
\label{sec:conclusion}
We presented a novel weakly supervised temporal action localization algorithm based on deep neural networks.
The classification is performed by evaluating a video-level representation given by a sparsely weighted mean of segment-level features where the sparse coefficients are learned with a sparsity loss in our deep neural network.
For weakly supervised temporal action localization, one-dimensional action proposals are extracted from which proposals relevant to target classes are selected to identify the time intervals of actions.
Our proposed approach achieved state-of-the-art performance on the THUMOS14 dataset, and we reported weakly supervised temporal action localization results on the ActivityNet1.3 dataset for the first time.

\vspace{-0.2cm}
\paragraph*{Acknowledgment}
We thank David Ross and Sudheendra Vijayanarasimhan at Google for providing the I3D features. 
This work is partly supported by the Korean ICT R\&D program of MSIP/IITP [2017-0-01780, 2016-0-00563].
 
{\small
\bibliographystyle{ieee}
\bibliography{cvpr2018_action}

\begin{thebibliography}{10}\itemsep=-1pt

\bibitem{alwassel17action}
H.~Alwassel, F.~C. Heilbron, and B.~Ghanem.
\newblock Action search: Learning to search for human activities in untrimmed
  videos.
\newblock In {\em arXiv preprint arXiv:1706.04269}, 2017.

\bibitem{bilen16weakly}
H.~Bilen and A.~Vedaldi.
\newblock Weakly supervised deep detection networks.
\newblock In {\em CVPR}, 2016.

\bibitem{bojanowski14weakly}
P.~Bojanowski, R.~Lajugie, F.~Bach, I.~Laptev, J.~Ponce, C.~Schmid, and
  J.~Sivic.
\newblock Weakly supervised action labeling in videos under ordering
  constraints.
\newblock In {\em ECCV}, 2014.

\bibitem{buch17sst}
S.~Buch, V.~Escorcia, C.~Shen, B.~Ghanem, and J.~C. Niebles.
\newblock {SST:} single-stream temporal action proposals.
\newblock In {\em CVPR}, 2017.

\bibitem{carreira17quo}
J.~Carreira and A.~Zisserman.
\newblock Quo vadis, action recognition? a new model and the kinetics dataset.
\newblock In {\em CVPR}, 2017.

\bibitem{deng09imagenet}
J.~Deng, W.~Dong, R.~Socher, L.-J. Li, K.~Li, and L.~Fei-Fei.
\newblock {ImageNet:} a large-scale hierarchical image database.
\newblock In {\em CVPR}, 2009.

\bibitem{escorcia16daps}
V.~Escorcia, F.~C. Heilbron, J.~C. Niebles, , and B.~Ghanem.
\newblock {DAPs:} deep action proposals for action understanding.
\newblock In {\em ECCV}, 2016.

\bibitem{feichtenhofer16spatiotemporal}
C.~Feichtenhofer, A.~Pinz, and R.~P. Wildes.
\newblock Spatiotemporal residual networks for video action recognition.
\newblock In {\em NIPS}, 2016.

\bibitem{feichtenhofer17spatiotemporal}
C.~Feichtenhofer, A.~Pinz, and R.~P. Wildes.
\newblock Spatiotemporal multiplier networks for video action recognition.
\newblock In {\em CVPR}, 2017.

\bibitem{feichtenhofer16convolutional}
C.~Feichtenhofer, A.~Pinz, and A.~Zisserman.
\newblock Convolutional two-stream network fusion for video action recognition.
\newblock In {\em CVPR}, 2016.

\bibitem{girdhar17actionvlad}
R.~Girdhar, D.~Ramanan, A.~Gupta, J.~Sivic, and B.~Russell.
\newblock Actionvlad: Learning spatio-temporal aggregation for action
  classification.
\newblock In {\em CVPR}, 2017.

\bibitem{gkioxari15finding}
G.~Gkioxari and J.~Malik.
\newblock Finding action tubes.
\newblock In {\em CVPR}, 2015.

\bibitem{gu17ava}
C.~Gu, C.~Sun, S.~Vijayanarasimhan, C.~Pantofaru, D.~A. Ross, G.~Toderici,
  Y.~Li, S.~Ricco, R.~Sukthankar, C.~Schmid, and J.~Malik.
\newblock {AVA}: A video dataset of spatio-temporally localized atomic visual
  actions.
\newblock In {\em arXiv:1705.08421}, 2017.

\bibitem{heilbron15activitynet}
F.~C. Heilbron, V.~Escorcia, B.~Ghanem, and J.~C. Niebles.
\newblock {ActivityNet:} a large-scale video benchmark for human activity
  understanding.
\newblock In {\em CVPR}, 2015.

\bibitem{heilbron16fast}
F.~C. Heilbron, J.~C. Niebles, and B.~Ghanem.
\newblock Fast temporal activity proposals for efficient detection of human
  actions in untrimmed videos.
\newblock In {\em CVPR}, 2016.

\bibitem{huang16connectionist}
D.-A. Huang, L.~Fei-Fei, and J.~C. Niebles.
\newblock Connectionist temporal modeling for weakly supervised action
  labeling.
\newblock In {\em ECCV}, 2016.

\bibitem{jiang14thumos}
Y.-G. Jiang, J.~Liu, A.~R. Zamir, G.~Toderici, I.~Laptev, M.~Shah, and
  R.~Sukthankar.
\newblock {THUMOS} challenge: Action recognition with a large number of
  classes, 2014.

\bibitem{karpathy14large}
A.~Karpathy, G.~Toderici, S.~Shetty, T.~Leung, R.~Sukthankar, and L.~Fei-Fei.
\newblock Large-scale video classification with convolutional neural networks.
\newblock In {\em CVPR}, 2014.

\bibitem{kay2017kinetics}
W.~Kay, J.~Carreira, K.~Simonyan, B.~Zhang, C.~Hillier, S.~Vijayanarasimhan,
  F.~Viola, T.~Green, T.~Back, P.~Natsev, et~al.
\newblock The kinetics human action video dataset.
\newblock {\em arXiv preprint arXiv:1705.06950}, 2017.

\bibitem{kuehne11hmdb}
H.~Kuehne, H.~Jhuang, E.~Garrote, T.~Poggio, and T.~Serre.
\newblock {HMDB:} a large video database for human motion recognition.
\newblock In {\em ICCV}, 2011.

\bibitem{laptive05on}
I.~Laptev.
\newblock On space-time interest points.
\newblock {\em IJCV}, 64(2-3):107--123, 2005.

\bibitem{ma16learning}
S.~Ma, L.~Sigal, and S.~Sclaroff.
\newblock Learning activity progression in lstms for activity detection and
  early detection.
\newblock In {\em CVPR}, 2016.

\bibitem{montes16temporal}
A.~Montes, A.~Salvador, S.~Pascual, and X.~Giro-i Nieto.
\newblock Temporal activity detection in untrimmed videos with recurrent neural
  networks.
\newblock In {\em 1st NIPS Workshop on Large Scale Computer Vision Systems
  (LSCVS)}, 2016.

\bibitem{richard16temporal}
A.~Richard and J.~Gall.
\newblock Temporal action detection using a statistical language model.
\newblock In {\em CVPR}, 2016.

\bibitem{rechard17weakly}
A.~Richard, H.~Kuehne, and J.~Gall.
\newblock Weakly supervised action learning with {RNN} based fine-to-coarse
  modeling.
\newblock In {\em CVPR}, 2017.

\bibitem{shi17learning}
Y.~Shi, Y.~Tian, Y.~Wang, W.~Zeng, and T.~Huang.
\newblock Learning long-term dependencies for action recognition with a
  biologically-inspired deep network.
\newblock In {\em ICCV}, 2017.

\bibitem{shou17cdc}
Z.~Shou, J.~Chan, A.~Zareian, K.~Miyazawa, and S.-F. Chang.
\newblock {CDC:} convolutional-de-convolutional networks for precise temporal
  action localization in untrimmed videos.
\newblock {\em CVPR}, 2017.

\bibitem{shou16temporal}
Z.~Shou, D.~Wang, and S.-F. Chang.
\newblock Temporal action localization in untrimmed videos via multi-stage
  cnns.
\newblock In {\em CVPR}, 2016.

\bibitem{simonyan14two}
K.~Simonyan and A.~Zisserman.
\newblock Two-stream convolutional networks for action recognition in videos.
\newblock In {\em NIPS}, 2014.

\bibitem{singh16multi}
B.~Singh, T.~K. Marks, M.~Jones, O.~Tuzel, and M.~Shao.
\newblock A multi-stream bi-directional recurrent neural network for
  fine-grained action detection.
\newblock In {\em CVPR}, 2016.

\bibitem{singh16untrimmed}
G.~Singh and F.~Cuzzolin.
\newblock Untrimmed video classification for activity detection: submission to
  {ActivityNet} challenge.
\newblock {\em arXiv preprint arXiv:1607.01979}, 2016.

\bibitem{singh17hide}
K.~K. Singh and Y.~J. Lee.
\newblock Hide-and-seek: Forcing a network to be meticulous for
  weakly-supervised object and action localization.
\newblock In {\em ICCV}, 2017.

\bibitem{soomro15action}
K.~Soomro, H.~Idrees, and M.~Shah.
\newblock Action localization in videos through context walk.
\newblock In {\em ICCV}, 2015.

\bibitem{soomro12ucf101}
K.~Soomro, A.~R. Zamir, and M.~Shah.
\newblock {UCF101:} a dataset of 101 human action classes from videos in the
  wild.
\newblock Technical Report CRCV-TR-12-01, University of Central Florida, 2012.

\bibitem{tran15learning}
D.~Tran, L.~D. Bourdev, R.~Fergus, L.~Torresani, and M.~Paluri.
\newblock Learning spatiotemporal features with {3D} convolutional networks.
\newblock In {\em ICCV}, 2015.

\bibitem{wang13action}
H.~Wang and C.~Schmid.
\newblock Action recognition with improved trajectories.
\newblock In {\em ICCV}, 2013.

\bibitem{wang13motionlets}
L.~Wang, Y.~Qiao, and X.~Tang.
\newblock Motionlets: Mid-level 3d parts for human motion recognition.
\newblock In {\em CVPR}, 2013.

\bibitem{wang16actioness}
L.~Wang, Y.~Qiao, X.~Tang, and L.~V. Gool.
\newblock Actionness estimation using hybrid fully convolutional networks.
\newblock In {\em CVPR}, 2016.

\bibitem{wang17untrimmednets}
L.~Wang, Y.~Xiong, D.~Lin, and L.~van Gool.
\newblock Untrimmednets for weakly supervised action recognition and detection.
\newblock In {\em CVPR}, 2017.

\bibitem{wang16temporal}
L.~Wang, Y.~Xiong, Z.~Wang, Y.~Qiao, D.~Lin, X.~Tang, and L.~val Gool.
\newblock Temporal segment networks: Towards good practices for deep action
  recognition.
\newblock In {\em ECCV}, 2016.

\bibitem{wang16anet}
R.~Wang and D.~Tao.
\newblock {UTS at Activitynet} 2016.
\newblock {\em AcitivityNet Large Scale Activity Recognition Challenge}, 2016.

\bibitem{wang17spatiotemporal}
Y.~Wang, M.~Long, J.~Wang, and P.~S. Yu.
\newblock Spatiotemporal pyramid network for video action recognition.
\newblock In {\em CVPR}, 2017.

\bibitem{wedel09improved}
A.~Wedel, T.~Pock, C.~Zach, H.~Bischof, and D.~Cremers.
\newblock {\em An Improved Algorithm for TV-$L^1$ Optical Flow}.
\newblock Statistical and geometrical approaches to visual motion analysis.
  Springer, 2009.

\bibitem{xiong17pursuit}
Y.~Xiong, Y.~Zhao, L.~Wang, D.~Lin, and X.~Tang.
\newblock A pursuit of temporal accuracy in general activity detection.
\newblock {\em arXiv preprint arXiv:1703.02716}, 2017.

\bibitem{xu17r}
H.~Xu, A.~Das, and K.~Saenko.
\newblock {R-C3D:} region convolutional 3d network for temporal activity
  detection.
\newblock In {\em ICCV}, 2017.

\bibitem{yeung16end}
S.~Yeung, O.~Russakovsky, G.~Mori, and L.~Fei-Fei.
\newblock End-to-end learning of action detection from frame glimpses in
  videos.
\newblock In {\em CVPR}, 2016.

\bibitem{yuan16temporal}
J.~Yuan, B.~Ni, X.~Yang, and A.~A. Kassim.
\newblock Temporal action localization with pyramid of score distribution
  features.
\newblock In {\em CVPR}, 2016.

\bibitem{yuan17temporal}
Z.~Yuan, J.~C. Stroud, T.~Lu, and J.~Deng.
\newblock Temporal action localization by structured maximal sums.
\newblock In {\em CVPR}, 2017.

\bibitem{zhao17temporal}
Y.~Zhao, Y.~Xiong, L.~Wang, Z.~Wu, X.~Tang, and D.~Lin.
\newblock Temporal action detection with structured segment networks.
\newblock In {\em ICCV}, 2017.

\bibitem{zhou16learning}
B.~Zhou, A.~Khosla, A.~Lapedriza, A.~Oliva, and A.~Torralba.
\newblock Learning deep features for discriminative localization.
\newblock In {\em CVPR}, 2016.

\end{thebibliography}
}

\end{document}